\documentclass[runningheads]{llncs}
\usepackage[T1]{fontenc}       
\usepackage{lmodern}           
\usepackage{textcomp}          

\usepackage{amsmath}
\usepackage{comment}
\usepackage{subfigure}
\usepackage{graphicx}
\usepackage{color}
\usepackage{xcolor}
\usepackage{amssymb}
\usepackage{cite}

\usepackage{xspace}
\usepackage{paralist}
\usepackage{hyperref}
\usepackage{pgfplots}
\usepackage{booktabs}    
\usepackage{multirow}    
\usepackage{wrapfig}  
\usepackage{microtype} 
\pgfplotsset{compat=1.18}

\newcommand{\plr}{\text{plr}\xspace{}}
\newcommand{\mysubsection}[1]{\medskip\noindent\textbf{#1}}
\DeclareMathOperator*{\argmmax}{arg\,max}

\hypersetup{
    linkcolor=blue,     
    citecolor=green,    
    urlcolor=magenta    
}

\begin{document}

\title{Statistical Runtime Verification for LLMs via Robustness Estimation}
\titlerunning{Statistical Runtime Verification for LLMs via Robustness Estimation}
\author{Natan Levy\textsuperscript{*} \and 
Adiel Ashrov\textsuperscript{*} \and
Guy Katz}

\authorrunning{N. Levy et al.}

\institute{The Hebrew University of Jerusalem, Jerusalem, Israel
\email{\{natan.levy1,adiel.ashrov,g.katz\}@mail.huji.ac.il}}

\maketitle              

\begin{abstract}
Adversarial robustness verification is essential for ensuring the safe deployment of Large Language Models (LLMs) in runtime-critical applications. However, formal verification techniques remain computationally infeasible for modern LLMs due to their exponential runtime and white-box access requirements. This paper presents a case study adapting and extending the RoMA statistical verification framework to assess its feasibility as an online runtime robustness monitor for LLMs in black-box deployment settings. Our adaptation of RoMA analyzes confidence score distributions under semantic perturbations to provide quantitative robustness assessments with statistically validated bounds. Our empirical validation against formal verification baselines demonstrates that RoMA achieves comparable accuracy (within 1\% deviation), and reduces verification times from hours to minutes. We evaluate this framework across semantic, categorial, and orthographic perturbation domains. Our results demonstrate RoMA's effectiveness for robustness monitoring in operational LLM deployments. These findings point to RoMA as a potentially scalable alternative when formal methods are infeasible, with promising implications for runtime verification in LLM-based systems.

\keywords{LLM safety \and Neural Network Verification \and LLM verification \and Robustness}
\end{abstract}

\renewcommand{\thefootnote}{*}
\footnotetext{Equal contribution}
\renewcommand{\thefootnote}{\arabic{footnote}}

\section{Introduction}

\emph{Large Language Models (LLMs)} such as GPT, BERT, and LLaMA~\cite{To23,RoTs2023,JaChLeTo18} operate over sequences of embedded tokens and obtain state-of-the-art results across diverse domains, demonstrating unprecedented capabilities in natural language understanding, reasoning, and generation tasks~\cite{gpt4,Ro23,To23}. This success has driven their rapid deployment across virtually every sector, from medicine~\cite{HaLeWi22} and education~\cite{LeDiLeWi25} to scientific research~\cite{ElLiBa24} and creative industries~\cite{Zh25}. LLMs are increasingly being integrated into safety-critical domains such as autonomous systems~\cite{HuSaMaGeCh24}, legal decision-making ~\cite{ChXiFeCh24}, and healthcare~\cite{HaLeWi22}, where their deployment will only continue to expand. However, as LLMs become widespread in applications where failures can lead to severe consequences, ensuring these powerful models are safe becomes crucial.

LLMs, like their Deep Neural Network (DNN) predecessors, are highly vulnerable to \emph{adversarial perturbations}: subtle modifications to input data that can cause incorrect outputs~\cite{BeJnCn24,SuBeXuMcSa19}. In contrast to adversarial attacks in computer vision, these perturbations in natural language are often coherent and contextually meaningful~\cite{QiLiZhGoZeZhWu25}, making them particularly challenging to detect using traditional validation methods~\cite{MoLiYoGrJiQi20}. This fragility poses significant risks in safety-critical systems, highlighting the need for runtime verification tools capable of monitoring LLM behavior during real-world deployment.

Formal verification methods~\cite{KaBaDiJuKo21,BrBaLi23} provide strong theoretical guarantees but scale poorly with large models, making them impractical for billion-parameter LLMs~\cite{KaHuIbJuLaLiShThWuZeDiKoBa19,WaZhXuLiJaHsKo21}. Statistical methods~\cite{WeRaTeKu19,HuHuHuPe21,CoRoKo19} offer better scalability but often rely on assumptions like Lipschitz continuity or Gaussian distributions, which are frequently violated in transformer-based models~\cite{KiPaMn21}. While several runtime verification techniques have been developed for neural networks, they are unsuitable for LLMs: Activation-based monitors~\cite{HaKrRiScVo24,HeWuBe24} require white-box access to model internals, and dynamic reachability methods like POLAR-Express~\cite{YaZhSiWaHuZh24} assume explicit system dynamics incompatible with natural language. These limitations highlight the critical gap in runtime verification capabilities for modern language models.

To address the gap in runtime verification for LLMs, we propose adapting the \emph{Robustness Measurement and Assessment (RoMA)} method~\cite{LeKa21} as a statistical framework for real-time robustness monitoring in operational environments. RoMA operates as a black-box framework, requiring no access to model internals, and efficiently handles high-dimensional inputs while empirically validating its statistical assumptions. These properties make RoMA particularly well-suited to overcome the limitations of existing verification approaches that fail to scale to modern LLMs. While RoMA has been used in the original study for vision classification tasks on small-scale networks, it has not previously been tested or adapted for LLMs, which is a novel contribution of this work.

This work presents a case study demonstrating the adaptation and extension of RoMA from offline image verification to online runtime monitoring for LLMs in operational environments.
We perform an empirical evaluation of BERT-based~\cite{JaChLeTo18} sentiment classifiers on the SST-2 dataset~\cite{SoPeWuChMaNgPo13}, examining three core robustness dimensions:
\begin{inparaenum}[(i)]
\item \emph {embedding robustness}, which measures how sensitive the model is to semantic perturbations in the Word2Vec~\cite{MiChCoDe13} embedding space;
\item \emph{categorial robustness}, measuring systematic performance differences across sentiment classes; and
\item \emph{orthographic robustness}, measuring model tolerance to typographical errors in real-world text.
\end{inparaenum} 
Our experimental results demonstrate that our adapted RoMA framework achieves computational efficiency suitable for runtime deployment: with 50\% of SST-2 sentences were processed within 15 minutes, with full evaluation completed in under 36 minutes. Our robustness assessment reveals that optimally trained models maintain 97.18\% robustness under semantic perturbations. Categorial analysis indicates systematic robustness differences of up to 1.5\% across sentiment classes. Additionally, orthographic analysis shows 94.44\% robustness under typographical errors. Taken together, these findings suggest that our adapted RoMA framework offers a scalable approach for runtime verification in LLMs.

An important consideration for statistical verification frameworks is whether they can deliver sufficiently accurate robustness estimates for practical deployment. To address this question, we empirically validate RoMA's accuracy against the \emph{Exact Count} formal verification algorithm~\cite{MaCoCiFa23}, which provides precise robustness measurements. We conduct systematic experiments across synthetic neural network models and the ACAS Xu safety-critical aviation benchmark~\cite{JuKoOw19,KaBaDiJuKo17} that appeared in the original Exact Count study. Our evaluation demonstrates that RoMA estimates robustness within 1\% deviation from Exact Count's results while reducing verification time from hours to minutes. This efficiency suggests that statistical frameworks like RoMA could be suitable for runtime monitoring, potentially helping to narrow the gap between theoretical robustness guarantees and practical operational constraints, though broader validation is needed to fully assess their applicability.

To summarize, our contributions include: 
\begin{inparaenum}[(i)]
\item adapting and extending RoMA from offline CNN verification to an online black-box runtime monitor for LLMs,
\item empirically validating its accuracy against formal verification baselines, showing comparable accuracy with reduced computational requirements, and 
\item demonstrating its application across embedding, categorial, and orthographic perturbation domains relevant to NLP deployments.
\end{inparaenum}

The rest of the paper is organized as follows. In Section
\ref{sec:background}, we provide the necessary background to contextualize our work. In Section \ref{sec:Related_Work} we review the related work to this paper. Building on this foundation, Section
\ref{sec:Method} introduces our proposed framework, detailing the
methodology for assessing LLM robustness. In Section
\ref{sec:evaluation}, we present our experimental setup, evaluation
metrics, and empirical findings, offering insights into the robustness
profiles of widely-used LLMs.  
Finally, in Section \ref{sec:conclusion}, we summarize our contributions and outline future research aimed at enhancing the reliability and trustworthiness of language models in real-world applications.

\section{Background}
\label{sec:background}

\mysubsection{DNNs, Adversarial Perturbations, and Robustness.}
\label{Background_DNNS_Robustness}
A DNN $N$ is defined as a function
$N: \mathbb{R}^n \rightarrow \mathbb{R}^m$ that maps an input vector
$\vec{x} \in \mathbb{R}^n$ to an output vector
$\vec{y} \in \mathbb{R}^m$. In this work, we focus on classification
networks, where an input $\vec{x}$ is classified as label $l$ when
$\argmmax(N(\vec{x}))=l$. In such networks, the final layer is usually a \emph{softmax layer}, whose outputs are commonly interpreted as \emph{confidence scores}.

In real-world settings, models are often exposed to \emph{adversarial
  perturbations}: input modifications that cause misclassification
yet remain imperceptible to human
operators~\cite{CaKaBaDi18,GoShSz15}. This vulnerability is
particularly problematic for runtime-critical systems, which require
mechanisms to monitor model behavior during operation and detect such
adversarial perturbations.

\emph{Local robustness} quantifies a network's resilience within some
input bounds \cite{KaHuIbJuLaLiShThWuZeDiKoBa19}:
\begin{definition}
	\label{definition1}
	A DNN $N$ is $\epsilon$-locally-robust at input point $\vec{x_0}$ if and only if
	\[
	\forall \vec{x}.
	\displaystyle || \vec{x} -\vec{x_0} ||_{\infty} \le \epsilon 
	\Rightarrow \argmmax(N(\vec{x})) = \argmmax(N(\vec{x_0})) 
	\]
\end{definition}

Intuitively, Definition~\ref{definition1} specifies that a DNN is
locally robust if, for all input vectors $\vec{x}$ within an
$\epsilon$-ball centered at a fixed input vector $\vec{x_0}$, the
network assigns the same label to $\vec{x}$ as it does to $\vec{x_0}$.
Verifying local robustness is computationally intractable for large
networks due to its NP-complete nature~\cite{KaBaDiJuKo21}. As a
result, runtime verification must often rely on \emph{probabilistic}
estimates of robustness.

\begin{definition}
	\label{definition2}
	The
	probabilistic-local-robustness (plr)
	score of a DNN $N$ at input point $\vec{x_0}$, abbreviated
	$\plr{}_{\epsilon}(N,\vec{x_0})$, is defined as:
	\begin{align*}
		\plr{}_{\epsilon}&(N,\vec{x_0})
		\triangleq 
		P_{x:  \lVert \vec{x} -\vec{x_0} \rVert_\infty \le \epsilon}
		(\argmmax(N(\vec{x})) = 
		\argmmax(N(\vec{x_0})))
	\end{align*}
\end{definition}

The $\plr{}$ score quantifies the likelihood that predictions remain unchanged within an $\epsilon$-ball and is particularly suitable for runtime certification under uncertainty~\cite{LaNi11}. Approximating $\plr{}$ efficiently enables real-time assessment of robustness in black-box settings~\cite{HuHuHuPe21,CoRoKo19,WeRaTeKu19}. This probabilistic measure aligns with certification standards requiring quantitative failure probability assessments, such as ARP 4754 guidelines~\cite{LaNi11}, while remaining computationally feasible for runtime evaluation.

\mysubsection{Statistical Verification Under Runtime Constraints.}

Statistical verification methods hold significant potential for addressing the scalability challenges that limit formal verification approaches. Here, we set out to examine this possibility by systematically evaluating statistical verification for LLM runtime monitoring.

The first approach involves methods that rely on \emph{Gaussian distributional assumptions}, such as randomized smoothing techniques~\cite{CoRoKo19}. However, these assumptions are frequently violated in transformer-based models~\cite{GuPlSuWe17}. The second approach is \emph{importance sampling} techniques such as~\cite{WeRaTeKu19}, but these can exhibit sensitivity to outlier samples, resulting in erratic monitoring behavior. Several approaches, such as Lipschitz-margin training\cite{TsSaSu18} and spectral norm regularization~\cite{YoMi17}, assume \emph{Lipschitz-continuity}, but these constants may not be well-defined for contemporary transformer architectures~\cite{KiPaMn21}. Finally, we select RoMA~\cite{LeKa21} as our statistical verification framework for LLM runtime monitoring because it addresses these limitations through several distinguishing characteristics.

RoMA~\cite{LeKa21} and gRoMA~\cite{LeYeKa23} are black-box robustness estimation frameworks designed to evaluate the robustness of DNNs. These methods analyze confidence scores from thousands of uniformly-sampled perturbations around an input point, employing Anderson-Darling goodness-of-fit~\cite{An11} testing to validate normal distribution assumptions. When necessary, they applies the Box-Cox power transformation~\cite{BoCo82} to achieve normality, enabling reliable probabilistic analysis. Specifically, RoMA focuses on the second-highest confidence score across all classes (the runner-up class), which reflects the margin between the predicted label and the nearest alternative. By analyzing the distribution of these scores across perturbations, RoMA quantifies how close the model is to misclassification, providing a sensitive measure of local robustness tied to Definition~\ref{definition2}. This approach enables probabilistic assessment of model reliability under the input variations encountered during operational deployment.

RoMA's key runtime advantages include no white-box access requirements, making it compatible with proprietary model deployments, empirically validated statistical assumptions that ensure distributional validity, and consistent computational overhead with linear scaling that enables predictable runtime resource consumption. Although RoMA was originally evaluated on vision tasks (CIFAR-10), it has not yet been extended to the unique challenges of high-dimensional NLP tasks. This gap is addressed in our work, where we apply the RoMA framework to LLMs, demonstrating its effectiveness in this new domain.

\section{Related work}
\label{sec:Related_Work}

\mysubsection{Robustness to Text Perturbations in Language Models.}
The robustness of language models to text perturbations is a vital research area, particularly as LLMs are increasingly deployed in real-world applications. Jin et al.~\cite{JiJiZhSz20} explored BERT’s vulnerability to adversarial attacks with \emph{TextFooler}, a method that generates adversarial examples through synonym replacement, revealing that even advanced models can be misled by subtle changes. In addition, Singh et al.~\cite{SiSiVa24} conducted a comprehensive analysis of LLM robustness to systematic text perturbations across different architectures and tasks, demonstrating that model behavior can be highly sensitive to small input changes. Finally, Romero-Alvarado et al.~\cite{RoHeMa24} investigated language models' resilience to various perturbation types, revealing systematic brittleness patterns across different input categories. While these studies provide valuable insights into LLM vulnerabilities through offline analysis, RoMA differentiates itself by enabling black-box statistical verification with quantitative robustness bounds designed for continuous monitoring during operational deployment.

\mysubsection{DNN Enable Monitor (DEM).} A closely related approach is \emph{DNN Enable Monitor (DEM)}~\cite{KaLeReYe24}, which provides output-centric, black-box certification for DNNs in safety-critical aerospace settings. While both DEM and our method rely on perturbation-based analysis without requiring model internals, they differ in scope and methodology. DEM applies hypothesis testing to detect adversarial inputs in image classifiers based on label consistency, producing binary accept/reject outcomes. Additionally, DEM requires a lengthy and sensitive calibration process before deployment, whereas our RoMA framework operates without any calibration requirements. Our RoMA-based framework targets LLMs and computes continuous robustness scores via distributional analysis of runner-up confidence margins under semantic, categorial, and orthographic perturbations. This enables more expressive and fine-grained monitoring suited to natural language domains.

\mysubsection{Runtime Monitoring for Neural Networks.} Runtime monitoring techniques for neural networks follow several paradigms. \emph{POLAR-Express}~\cite{YaZhSiWaHuZh24} performs online reachability analysis in NN-controlled systems, enabling dynamic controller switching upon detecting unsafe states. The combined \emph{Gaussian and Outside-the-Box monitor}~\cite{HaKrRiScVo24} detects \emph{out-of-distribution (OOD)} inputs via activation-based analysis, blending neuron-wise Gaussian modeling with clustering. Similarly, the \emph{box-based monitor for YOLO}~\cite{HeWuBe24} constructs hyper-rectangular activation zones to identify OOD behavior at runtime. While effective, these approaches rely on binary decisions and require access to internal activations~\cite{HaKrRiScVo24,HeWuBe24} or explicit dynamics~\cite{YaZhSiWaHuZh24}. In contrast, RoMA’s black-box methodology supports scalable, model-agnostic monitoring for LLMs, providing probabilistic robustness estimates that capture nuanced behavior under realistic input shifts.

\section{Method: Statistical Distribution Analysis for LLM Classification Resilience}
\label{sec:Method}

The RoMA framework is a statistical verification technique originally developed to assess the robustness of image classifiers. RoMA operates as a black-box method, meaning it requires no access to internal model weights or gradients. Instead, it evaluates robustness by sampling perturbations around a given input and analyzing their effect on the model's confidence scores. A key innovation of RoMA is its focus on the \emph{runner-up confidence score}—the second-highest class probability—across perturbed inputs. This score serves as a proxy for how close the model is to changing its prediction. For example, consider a classifier that assigns \texttt{cat} to an image with 92\% confidence and \texttt{dog} with 8\%. If slight perturbations cause the runner-up score (\texttt{dog}) to rise significantly, this indicates an unstable prediction. RoMA collects thousands of such perturbed inputs, tests whether the runner-up scores follow a normal distribution using the Anderson-Darling test~\cite{An11}, and applies the Box-Cox transformation~\cite{BoCo82} if needed to enforce normality. This enables robust estimation of the probability that random perturbations will flip the classification, providing a quantitative measure of local robustness (plr).

While RoMA was originally designed for image classifiers, extending it to LLMs introduces several significant challenges. First, unlike images, where perturbations involve continuous pixel value changes, language inputs consist of discrete tokens embedded in high-dimensional semantic spaces. Text perturbations must preserve syntactic validity and semantic meaning, and arbitrary token replacements can easily produce nonsensical inputs. Second, in practical deployment settings, LLMs are typically accessed as black-box services via APIs, providing only output probabilities without access to internal embeddings or model parameters. Third, the distributional assumptions underpinning RoMA, particularly the normality of runner-up confidence scores, are not guaranteed to hold for natural language data, which exhibits complex statistical patterns distinct from image data. These challenges necessitate fundamental adaptations in how perturbations are generated, how model responses are analyzed, and how statistical validity is ensured for NLP applications.

Word embeddings play a central role in natural language processing by mapping discrete words to continuous high-dimensional vectors that capture semantic similarity. In this space, words with related meanings are located near one another, enabling quantitative reasoning over language. Embedding models such as Word2Vec~\cite{MiChCoDe13}, GloVe~\cite{PeSoMa14}, and contextualized embeddings from transformers like BERT~\cite{JaChLeTo18} allow us to compare words using cosine similarity, facilitating controlled semantic perturbations without compromising linguistic validity. In our adaptation, we use pre-trained Word2Vec embeddings as a lightweight and interpretable basis for generating meaning-preserving input variations. While Word2Vec was chosen for its simplicity and transparency, our framework is general and can accommodate other perturbation techniques, such as paraphrasing or syntactic restructuring, which may further enhance robustness evaluation.

Building on this embedding-based representation of semantic similarity, we adapt RoMA's perturbation strategy to the linguistic domain. Instead of injecting pixel-level noise as in vision tasks, we introduce semantically meaningful perturbations by replacing words with similar alternatives in the embedding space. Specifically, we use Word2Vec embeddings to identify replacement candidates whose cosine similarity to the original word exceeds a threshold of $1 - \epsilon$, where $\epsilon$ is derived from the original RoMA framework and controls the magnitude of allowed semantic drift. This constraint ensures that perturbed sentences remain semantically coherent while still exploring the model's decision boundaries. For example, with $\epsilon = 0.35$ (as used in our experiments), when perturbing the word ``good'' in the sentence \texttt{``This movie is really good''}, we might select ``great'' (word similarity to ``good'': 0.68) or ``excellent'' (word similarity to ``good'': 0.73), yielding variants such as \texttt{``This movie is really great''} and \texttt{``This movie is really excellent''}. By generating hundreds of such semantically constrained perturbations and analyzing the resulting confidence distributions, our adaptation enables RoMA's statistical framework to operate on LLMs while preserving the black-box deployment setting required for runtime verification.

Several perturbation strategies have been explored for evaluating LLM robustness, each with trade-offs that limit their applicability for runtime verification. One approach perturbs internal model embeddings directly by modifying hidden layer activations~\cite{MoJuHiYu18}, but this requires white-box access incompatible with most deployed systems. Another common strategy is character-level noise injection, simulating typographical errors by inserting or substituting letters~\cite{SiSiVa24}. While such perturbations may not always preserve full semantic coherence, they offer a lightweight and realistic proxy for input noise encountered in real-world deployments, and can still yield meaningful robustness estimates, as demonstrated in our orthographic robustness evaluation (See Section~\ref{sec:roma_Typo_llm}). Finally, semantic substitution, replacing words with meaning-preserving alternatives, offers the most promise for black-box runtime verification as it maintains linguistic validity while testing model stability. This approach, which we implement through controlled word embedding similarities, forms the foundation of our adaptation.

Our perturbation generation process systematically explores the semantic neighborhood around each input while maintaining linguistic validity. Given an input sentence, we first tokenize it and randomly select multiple word positions for perturbation, ensuring coverage across the entire sentence rather than concentrating changes in a single region. For each selected word, we query the Word2Vec~\cite{MiChCoDe13} embedding space to retrieve semantically similar candidates whose cosine similarity exceeds a threshold of $(1 - \epsilon)$. To preserve sentence quality, we filter out stopwords, proper nouns, and out-of-vocabulary terms that could compromise coherence. By sampling different combinations of word replacements across multiple positions, we generate up to 1,000 unique variants per sentence, sufficient for RoMA's statistical analysis while ensuring comprehensive exploration of the local semantic space. This structured approach balances semantic coverage with computational efficiency, enabling robust statistical estimation within runtime constraints.

For each perturbed variant generated from a given input sentence, we query the LLM sentiment classifier to obtain confidence scores for the positive and negative sentiment classes. Following the RoMA framework, we focus on the runner-up confidence score, which in binary sentiment analysis corresponds to the confidence score of the non-predicted class. For example, if the model predicts ``positive'' sentiment with 85\% confidence, the runner-up score is 15\% for ``negative''. A high runner-up score signals uncertainty and indicates that the model’s prediction could easily flip under slight perturbations. By aggregating runner-up scores across all perturbations (up to 1,000 per sentence), we construct an empirical distribution that captures classification stability in the local semantic neighborhood. To validate this distribution for statistical analysis, we apply the Anderson-Darling goodness-of-fit test~\cite{An11} to assess normality. When this assumption is violated, we apply the Box-Cox power transformation~\cite{BoCo82} to approximate normality and enable reliable probabilistic inference. This allows us to estimate the probability that a random semantic perturbation will cause the model to misclassify the sentiment, yielding a quantitative robustness score aligned with the probabilistic-local-robustness (plr) metric defined in Section~\ref{definition2}.

While RoMA was originally designed as an offline evaluation tool for image classifiers, extending it to online runtime LLM verification required significant methodological modifications. These adjustments include developing semantically-constrained text perturbations (replacing pixel noise), validating statistical assumptions for transformer confidence distributions, and addressing the strict computational constraints of continuous monitoring. Our framework operates entirely through black-box API queries, making it suitable for proprietary LLM deployments where internal model access is unavailable. Perturbation analysis is performed during inference gaps or alongside batched requests, allowing for continuous robustness monitoring without disrupting service. The linear scalability of perturbation generation and consistent processing times support predictable resource allocation, which is critical for deployment scenarios where latency must remain bounded. In summary, our work transforms an offline framework into an online runtime verification system, demonstrating the potential of statistical verification for practical LLM monitoring in operational environments.

\section{Evaluation}
\label{sec:evaluation}

Our experimental evaluation comprises two complementary phases designed to establish both the empirical validity and practical applicability of our runtime verification framework. We first conduct an empirical validation of RoMA's statistical methodology against Exact Count, a formal verification baseline, to demonstrate precision and reliability on established benchmarks where ground truth verification is computationally feasible. This validates RoMA's statistical approach and quantifies its accuracy relative to exhaustive formal methods.

Subsequently, we demonstrate the framework's significance by applying it to contemporary LLM verification scenarios that exceed the computational scope of formal methods. Our evaluation addresses three dimensions of LLM verification:
\begin{inparaenum}[(i)]
    \item embedding space robustness under semantic perturbations, 
    \item categorial performance consistency across classification boundaries, and
    \item orthographic resilience to typographical variations commonly encountered in operational deployments.
\end{inparaenum}
These evaluations collectively establish RoMA's capability to provide quantitative robustness guarantees for large-scale neural architectures while maintaining computational efficiency suitable for runtime monitoring.

All experiments were conducted on dedicated research infrastructure to ensure reproducible and controlled evaluation conditions. The baseline validation experiments (Section \ref{sec:roma_vs_exact_count}) utilized an AMD EPYC 7313 CPU with 128GB RAM and NVIDIA A10 GPU acceleration. The LLM verification experiments (Sections \ref{sec:roma_and_llm_embeddings}, \ref{sec:Categorial_Robustness}, and \ref{sec:roma_Typo_llm}) were performed using equivalent computational resources with NVIDIA A5000 GPU acceleration to accommodate the increased memory requirements of transformer-based architectures.
Complete implementation details, experimental configurations, and reproducibility materials are publicly available through our research repository~\cite{ourCode}, enabling independent validation and extension of our findings.

\subsection{Validating RoMA against Formal Verification}
\label{sec:roma_vs_exact_count}

Formal verification methods provide mathematical guarantees by proving the absence of adversarial perturbations within specified input regions. If adversarial perturbations exist, a formal verifier will usually return a concrete example that demonstrates this. These techniques rely on SMT solving, abstract interpretation, or reachability analysis~\cite{KaBaDiJuKo21,BrBaLi23}, and are implemented in tools such as Marabou~\cite{KaHuIbJuLaLiShThWuZeDiKoBa19}, Beta-CROWN~\cite{WaZhXuLiJaHsKo21}, and PyRAT~\cite{ArTeBu11}. However, to the best of our knowledge, such methods do not scale to the architectural complexity and size of modern LLMs, which include components like multi-head attention and large embedding layers. This motivates our use of smaller models, where exact robustness bounds remain tractable and can serve as a reliable reference.

Formal verification's primary strength is \emph{soundness}: once verified, a property holds universally. However, their practical use in runtime scenarios is limited due to the following reasons:
\begin{inparaenum}[(i)]
\item the verification problem is NP complete~\cite{KaBaDiJuKo21},
  making it challenging for verifiers to scale to modern LLMs;
\item verification techniques typically require white-box access to a
  model's parameters, which is often impossible; and
\item the high latency of these approaches typically renders them unsuitable for deployment-time constraints.
\end{inparaenum}

The \emph{Exact Count} algorithm~\cite{MaCoCiFa23} constitutes a formal methodology for the precise quantification of DNN safety violations through exhaustive domain analysis. This approach implements a systematic recursive partitioning strategy of the input space, decomposing it into regions that can be definitively classified according to their adherence to specified safety properties.

Exact Count computes the violation rate with mathematical precision, defined formally as the ratio of unsafe regions to the total input space volume. This metric provides an exact measure of the \emph{probability of encountering adversarial perturbations}, which is $1-\plr{}$ as defined in Definition \ref{definition2}. This establishes a connection between formal verification and probabilistic robustness assessment.

However, the computational complexity of Exact Count, which increases exponentially with input dimensionality and network size, fundamentally constrains its practical application. Even for relatively small networks like those in the ACAS Xu collision avoidance systems~\cite{OwPaMo19} (approximately 300 neurons), Exact Count becomes computationally intractable within practical time limits. Our experimental evaluation shows that Exact Count consistently timed out after 24 hours (See Table~\ref{tab:robustness-comparison}). This highlights the need for more scalable alternatives, such as RoMA, which was able to produce reliable estimates for the same models in under 16 minutes.

\emph{CountingProVe}~\cite{MaCoCiFa23} offers a scalable, randomized alternative by sampling subregions and bounding the violation rate statistically. While more tractable, it sacrifices precision and requires repeated solver queries that limit its applicability in real-time systems.

We acknowledge that the feedforward networks and ACAS Xu models used in our formal validation differ from transformer-based LLMs in architectural complexity, including the presence of multi-head attention, positional encodings, and high-dimensional embeddings. While this validation demonstrates the statistical reliability of RoMA’s methodology, it does not capture these LLM-specific characteristics. The LLM experiments in subsequent sections provide complementary evidence of RoMA’s effectiveness on transformer architectures, although direct comparison with formal verification remains infeasible for such large-scale models.

\mysubsection{Experimental Design.}
To establish the accuracy and reliability of our statistical verification framework, we conduct a comprehensive empirical validation against the Exact Count algorithm~\cite{MaCoCiFa23}, a formal verification method that computes mathematically precise probabilistic robustness (\plr{}) scores. While Exact Count provides definitive ground truth robustness measurements for small-scale neural networks, its exponential computational complexity fundamentally limits applicability to contemporary large-scale architectures. Nevertheless, it serves as an authoritative benchmark for evaluating the precision of RoMA's probabilistic approximations under rigorously controlled experimental conditions.

We implemented a complete reproduction of the Exact Count algorithm and conducted systematic experiments across two established benchmark suites from the original paper:
\begin{inparaenum}[(i)]
\item synthetic neural network models with varying architectural complexities, and
\item the ACAS Xu safety-critical aviation collision avoidance benchmark~\cite{KaBaDiJuKo21}, representing real-world verification challenges in autonomous systems.
\end{inparaenum}
This benchmark selection ensures comprehensive evaluation across both controlled synthetic scenarios and practical safety-critical applications, providing robust validation of RoMA's statistical methodology across diverse verification contexts.

\begin{figure}[t]
    \centering
    \begin{minipage}[t]{0.45\textwidth}
        \centering
        \includegraphics[page=1,
                    trim=50 210 20 235,
                    clip,
                    width=\textwidth]{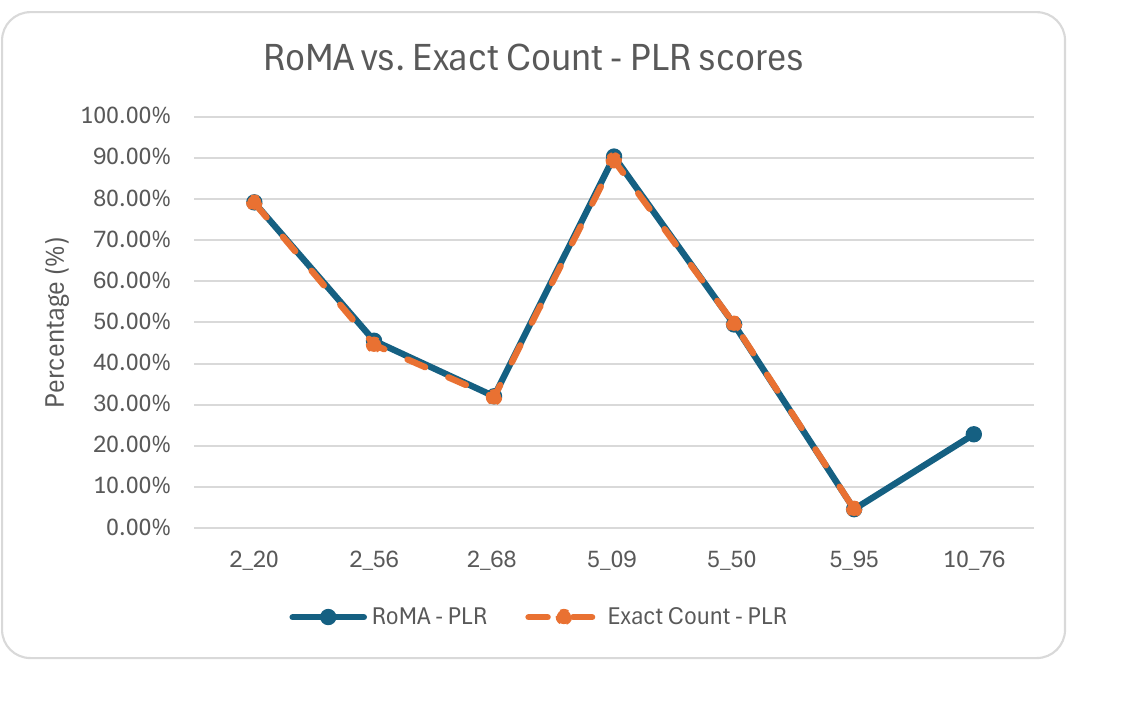}
        
        \caption{\emph{PLR} scores of RoMA and the Exact Count algorithm across benchmark models.}
        \label{fig:plr_roma_vs_exact_count}
    \end{minipage}
    \hfill
    \begin{minipage}[t]{0.45\textwidth}
        \centering
        \includegraphics[page=1,
                    trim=50 483 150 50,
                    clip,
                    height=0.75\textwidth, 
                    width=\textwidth,
                    keepaspectratio]{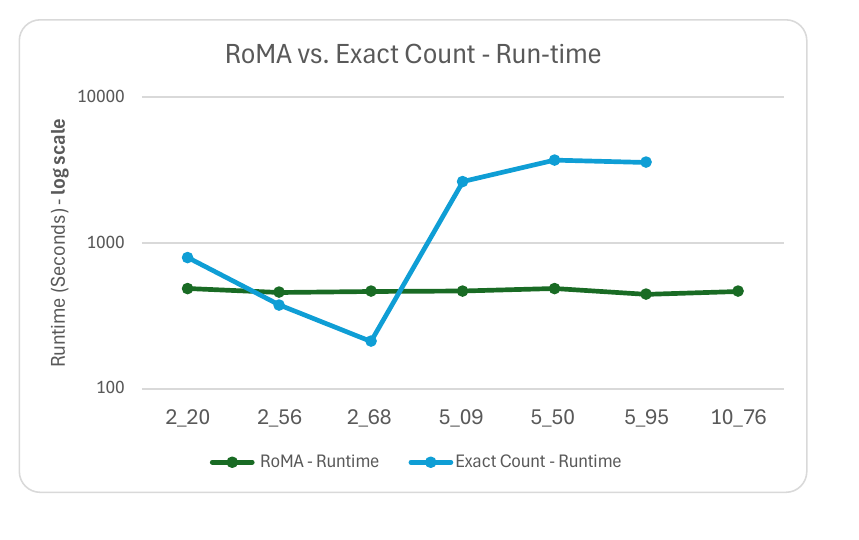}
        
        \caption{Runtime comparison between RoMA and the Exact Count algorithm.}
        \label{fig:runtime_roma_vs_exact_count}
    \end{minipage}
\end{figure}

\mysubsection{Results.}
The scalability advantages of RoMA become particularly pronounced for larger networks, including ACAS Xu benchmarks with property $\phi_2$, where Exact Count consistently terminates with a timeout after 24 hours of computation. In contrast, RoMA provides reliable and accurate robustness measurements in under 16 minutes for these identical verification challenges, demonstrating applicability to complex models that fundamentally exceed the computational reach of formal verification methods.
These results, summarized in Table~\ref{tab:robustness-comparison} and illustrated in Figs.~\ref{fig:plr_roma_vs_exact_count} and~\ref{fig:runtime_roma_vs_exact_count}, establish RoMA's capability to address the computational intractability barrier that prevents formal verification deployment in runtime scenarios.

\begin{table}[t]
    \centering
    \caption{RoMA vs. Exact Count --- Robustness Measurement}
    \begin{tabular}{l|ccc|cc}
        \toprule
        \multirow{2}{*}{Model} & 
        \multicolumn{3}{c|}{$Exact$ $Count$} & 
        \multicolumn{2}{c}{$RoMA$} \\
        \cmidrule(lr){2-4} \cmidrule(lr){5-6}
        & $Violation$ $Rate$ & $PLR$ & $Run$ $time$ & $PLR$ & $Run$ $time$ \\
        \midrule
        Model\_2\_20 & 20.88\% & 79.12\% & 794~sec & 79.24\% & 487~sec \\ 
        Model\_2\_56 & 55.44\% & 44.56\% & 374~sec & 45.56\% & 458~sec \\
        Model\_2\_68 & 68.20\% & 31.80\% & 211~sec & 32.06\% & 466~sec \\
        Model\_5\_09 & 10.60\% & 89.40\% & 2,636~sec & 90.36\% & 467~sec \\
        Model\_5\_50 & 50.33\% & 49.67\% & 3,696~sec & 49.52\% & 486~sec \\
        Model\_5\_95 & 95.35\% & 4.65\% & 3,561~sec & 4.59\% & 444~sec \\
        Model\_10\_76 & --- & --- & 24~hrs & 22.78\% & 465~sec \\
        \midrule
        $\phi_2$ ACAS Xu\_2.1 & --- & --- & 24~hrs & 99.18\% & 775~sec \\
        $\phi_2$ ACAS Xu\_2.3 & --- & --- & 24~hrs & 98.24\% & 638~sec \\
        $\phi_2$ ACAS Xu\_2.4 & --- & --- & 24~hrs & 98.96\% & 915~sec \\
        $\phi_2$ ACAS Xu\_2.5 & --- & --- & 24~hrs & 98.09\% & 679~sec \\
        $\phi_2$ ACAS Xu\_2.7 & --- & --- & 24~hrs & 97.21\% & 616~sec \\
        \bottomrule
    \end{tabular}
    \label{tab:robustness-comparison}
\end{table}

\mysubsection{Implications for Runtime Verification.}
The experimental validation demonstrates that RoMA can bridge the gap between theoretical verification guarantees and practical runtime applicability. The consistent sub-1\% accuracy achieved across diverse benchmark scenarios, combined with predictable computational overhead independent of model scale, establishes RoMA as a statistically accurate and operationally viable alternative to formal verification for large-scale neural network verification.
While these empirical results cannot provide universal mathematical guarantees across all possible network architectures, they offer compelling evidence of RoMA's precision and computational efficiency across representative verification scenarios. This validation extends statistical verification to contemporary LLM architectures.
The demonstrated reliability of RoMA's statistical methodology against formal verification baselines provides the necessary confidence to proceed with LLM verification applications, where ground truth formal verification is computationally infeasible but statistical reliability assessment remains critical for operational deployment.

\subsection{Statistical Verification of LLM Embedding Robustness}
\label{sec:roma_and_llm_embeddings}

\mysubsection{Experimental Design.}
To demonstrate the practical applicability of our statistical verification framework for contemporary language models, we conducted a comprehensive robustness assessment on fine-tuned BERT architectures~\cite{JaChLeTo18} using the SST-2 classification task from the GLUE benchmark~\cite{WaXuWaGa21}. This evaluation employs two distinct BERT-base-uncased variants~\cite{bert_base_uncased} (110M parameters) to examine how training optimization affects distributional robustness under semantic perturbations.
The GLUE dataset and BERT model have been widely used to evaluate generalization and robustness in NLP models, and have become the de-facto standard frameworks for such assessments. Our experimental design utilizes two model configurations representing different training strategies:
$M_{\text{best}}$, corresponding to the optimal performance checkpoint during training, and $M_{\text{final}}$, representing the final training iteration.

For each of the 1,821 test sentences in the SST-2 evaluation set, we applied our semantic perturbation methodology to generate up to 1,000 variations per input, subsequently analyzing the distributional properties of runner-up confidence scores to assess classification stability. The statistical validation process revealed that confidence score distributions satisfied normality assumptions in 81.60\% of cases for $M_{\text{best}}$ and 75.07\% for $M_{\text{final}}$, demonstrating the reliability of our distributional approach for the majority of inputs.
This high rate of distributional normality validates the fundamental statistical assumptions underlying our verification framework, providing confidence in the reliability of probabilistic robustness estimates for operational LLM deployment scenarios.
\begin{figure}[ht]
    \centering
    \includegraphics[page=1, trim=50 425 100 50, clip,width=0.9\textwidth]{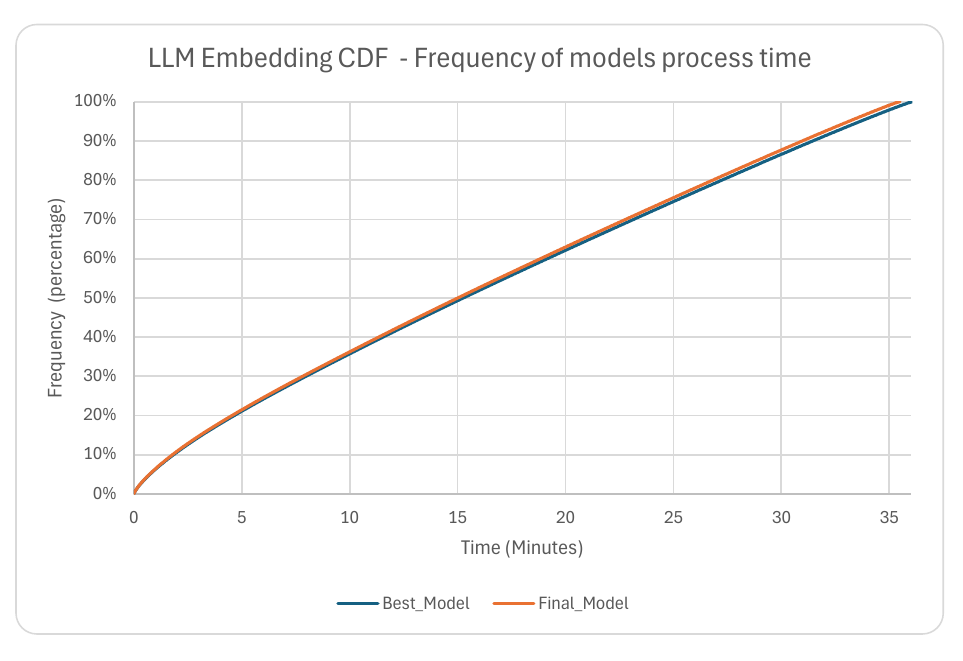}
    \caption{Cumulative distribution function (CDF) showing the percentage of SST-2 dataset instances processed over time by RoMA in the LLM embedding case study for the two models.}
    \label{fig:LLM_embeddings_CDF}
\end{figure}

\mysubsection{Results.}
We quantified embedding robustness as the percentage of semantically perturbed inputs maintaining classification confidence scores above the 0.50 threshold for correct sentiment classification. Lower confidence scores indicate reduced classification certainty and potential vulnerability to distributional shifts encountered during runtime operation.

The robustness performance results show 
97.18\%  robustness score for $M_{\text{best}}$, and 
96.60\% robustness score for $M_{\text{final}}$. The superior robustness exhibited by $M_{\text{best}}$ supports the hypothesis that optimization for classification performance may simultaneously enhance resilience to semantic perturbations, a finding which is consistent with previous work on neural network robustness~\cite{LeKa21}. This correlation suggests that performance-driven training optimization can contribute to improved distributional stability, with important implications for model selection in runtime-critical applications.

Beyond robustness quantification, we evaluated the computational efficiency of our statistical verification approach to determine its viability for large-scale LLM assessment in production environments. Figure~\ref{fig:LLM_embeddings_CDF} presents the \emph{Cumulative Distribution Function (CDF)} of RoMA processing times across the complete evaluation dataset. We observe that 50\% of the instances were processed within 15 minutes, and the entire dataset was evaluated in under 36 minutes.

\mysubsection{Implications for Runtime Verification.}
These results suggest that our statistical framework meets the performance demands of runtime monitoring in deployed LLM systems. By delivering timely robustness estimates based on semantically meaningful perturbations, our adaptation and extension of RoMA enables continuous assessment of classification stability during operation. This capability is essential for runtime verification pipelines that must monitor model behavior under distributional drift, without relying on white-box access or introducing latency that disrupts real-time service.

\subsection{Categorial Robustness}
\label{sec:Categorial_Robustness}

Prior work in computer vision has established that neural network robustness exhibits significant variation across distinct input categories~\cite{LeKa21}. To examine whether this categorial heterogeneity extends to natural language processing architectures, we conducted a systematic analysis of distributional robustness patterns across sentiment classification categories using our statistical verification framework.

We define \emph{categorial robustness} as the statistical measure of model resilience computed independently for each classification category, enabling identification of systematic vulnerabilities that may not be apparent in aggregate robustness assessments. For our binary sentiment analysis evaluation, we partitioned the SST-2 test dataset according to ground truth labels (positive and negative sentiment) and calculated category-specific robustness scores through our semantic perturbation methodology.

\mysubsection{Experimental Design.}
Categorial Analysis Protocol:
\begin{inparaenum}[(i)]
    \item Partition inputs by true classification labels to isolate category-specific behavior
    \item Apply the semantic perturbation framework independently within each category
    \item Compute distributional statistics for runner-up confidence scores per category
    \item Calculate category-specific robustness metrics using the 0.50 confidence threshold
    \item Analyze asymmetric patterns across classification boundaries
\end{inparaenum}

\begin{figure}
    \centering
    \includegraphics[page=1, trim=100
    500 200 75, width=0.75\textwidth]{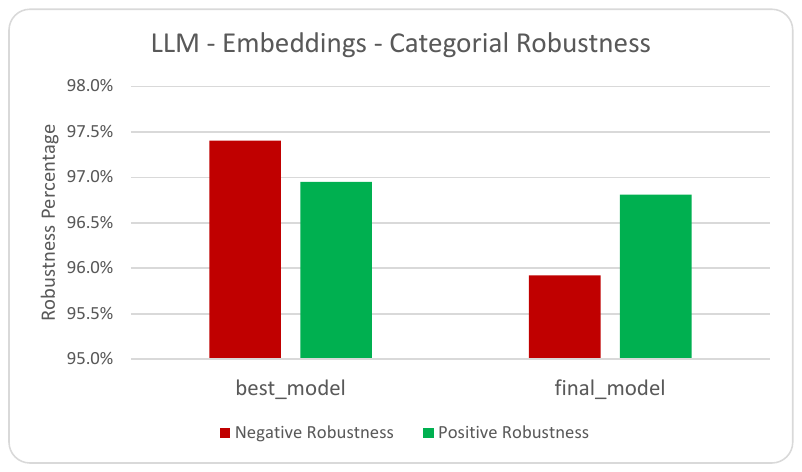}
    \caption{A comparison of categorial robustness between
      $M_{\text{best}}$ and $M_{\text{final}}$.}
    \label{fig:categorial_robustness_llm}
\end{figure}

\mysubsection{Results.} Our analysis reveals systematic variation in distributional robustness across categories, consistent with prior observations in computer vision architectures~\cite{LeKa21}. This suggests that asymmetries in categorial robustness may be an inherent property of neural networks. Figure~\ref{fig:categorial_robustness_llm} illustrates the categorial robustness across sentiment classes for both $M_{\text{best}}$ and $M_{\text{final}}$.

\mysubsection{Implications for Runtime Verification.}
These findings have important implications for runtime verification, as they highlight that model resilience is not uniformly distributed across classes. Runtime monitors must therefore account for class-conditional robustness profiles to ensure reliable behavior across all inputs. Integrating category-aware metrics into verification pipelines can enable early detection of systematic vulnerabilities and support targeted mitigation strategies in safety-critical NLP applications.

\subsection{Orthographic Perturbation Analysis for Runtime Input Validation}
\label{sec:roma_Typo_llm}

\mysubsection{Experimental Design.} To evaluate LLM resilience to typographical errors in operational environments, we implemented systematic character-level perturbations simulating common human typing mistakes. Specifically, each character in every word was systematically replaced with all possible alphabetic alternatives, generating a broad set of single-character substitution variants. While this method is simple by design, it provides comprehensive coverage of potential character-level noise in user input, such as accidental character substitutions (e.g., ``great'' → ``grebt''). We limited perturbations to alphabetic characters, avoiding modifications to whitespace and punctuation. We acknowledge that this process does not replicate empirical human error distributions, but rather serves as a lightweight proxy for orthographic noise. Incorporating more realistic perturbation models, such as those based on keyboard adjacency or observed typo patterns, remains a promising direction for future work.

\mysubsection{Results.}
Analysis of 500 SST-2 sentences with 500 character-level perturbations each revealed that runner-up confidence scores failed Anderson-Darling goodness-of-fit normality tests even after Box-Cox transformation. Despite non-normal distributions, our framework produced robustness estimates of 94.44\% for $M_{\text{best}}$ and 93.94\% for $M_{\text{final}}$.
To validate these estimates, we conducted an exhaustive evaluation across all 1,821 test sentences, yielding ground truth scores of 94.61\% and 93.84\% respectively. 

\mysubsection{Implications for Runtime Verification.}
This agreement between estimated and ground truth robustness scores (within 0.17\%), provides preliminary evidence that RoMA may remain effective even when distributional assumptions such as normality are violated. This potential resilience is relevant for runtime verification, where inputs are often subject to noise, typographical errors, or other irregularities. While further validation is needed, these results suggest that statistically grounded robustness assessments could remain informative under realistic deployment conditions.

\section{Conclusion and Future Work}
\label{sec:conclusion}

This paper presented a case study on adapting and extending the RoMA framework for runtime robustness assessment of LLM systems. We examine the feasibility of applying statistical verification techniques for continuous reliability monitoring in black-box settings, where white-box access is not available. The case study illustrates how RoMA could potentially support LLM robustness auditing under practical deployment constraints. Preliminary empirical comparisons with the Exact Count formal verification baseline indicate that RoMA can approximate robustness within sub-1\% error margins, while reducing computation time significantly. While these results are encouraging, they represent an initial step toward evaluating the role of statistical methods in runtime verification for large-scale models.

\mysubsection{Technical Contributions.}
This study presents an initial exploration into adapting and extending RoMA for runtime robustness assessment of LLMs. The proposed methodology incorporates the following components:
\begin{inparaenum}[(i)]
\item an adaptation of RoMA from offline evaluation to online monitoring for language models in black-box settings,
\item a semantic perturbation strategy based on word embedding transformations to examine distributional sensitivity,
\item a categorial robustness analysis aimed at identifying potential variation in resilience across sentiment classes, and
\item an orthographic perturbation evaluation designed to assess model behavior under character-level input noise.
\end{inparaenum}
These contributions form the basis for assessing runtime behavior under realistic perturbation domains, though further validation is needed to generalize beyond the specific case study explored here.

\mysubsection{Operational Impact.}
Our analysis suggests that robustness characteristics may vary across models and input categories, highlighting the potential value of adaptive monitoring strategies in practice. The statistical nature of the proposed methodology, which does not rely on internal model access, indicates that it may be applicable to a range of LLM deployments, including settings where models are accessed through black-box APIs. Initial findings on computational efficiency point toward the feasibility of integrating such methods into runtime environments, where verification must be performed under time and resource constraints. Further investigation is needed to confirm these observations across a broader range of applications and model types.

\mysubsection{Future Directions.}
Several directions remain for further exploration. One avenue is to extend the framework to additional supervised learning domains, such as speech recognition, to examine its applicability beyond text-based tasks. Another is to adapt the approach for reinforcement learning, which poses unique challenges for runtime verification. In addition, incorporating more diverse semantic perturbation techniques, such as paraphrasing, sentence restructuring, or syntactic transformations, could further enrich the robustness evaluation beyond simple synonym substitution with Word2Vec. Another promising direction is to perform empirical comparisons with established robustness benchmarks and attack frameworks, such as TextFooler, Adversarial GLUE, and other recent evaluations~\cite{DoLuTuJi21, WaZh24, JoJiRaLi20}, which would provide valuable insights into the scalability and competitiveness of our approach. Lastly, while this study focused on encoder-only architectures (specifically BERT), future work should evaluate the applicability of the proposed framework to decoder-only models~\cite{openai2025chatgpt} and encoder-decoder architectures~\cite{RaShRoLeNaMaZhLiLi23}. These directions may help assess the generalizability of the framework and identify domain-specific considerations for broader deployment.

\mysubsection{Runtime Verification Contribution.}
This study explores the use of a statistical methodology for monitoring learning-enabled components in operational settings where formal verification techniques may be impractical. The proposed framework offers an initial step toward enabling continuous reliability assessment in large-scale neural architectures, particularly in scenarios where white-box access is unavailable. While preliminary, the ability to estimate classification resilience under perturbations may contribute to the development of runtime assurance strategies for safety-critical applications that rely on LLMs.

\section*{\uppercase{acknowledgments}}
We would like to thank Davide Corsi for his 
assistance in this project and his insightful 
comments. 
This work was partially 
funded by the European Union (RobustifAI project, ID 101212818). Views and opinions expressed are however those of the
author(s) only and do not necessarily reflect those of the European Union or the European
Health and Digital Executive Agency (HADEA). Neither the European Union nor the granting
authority can be held responsible for them.  Additionally, this work was partially 
supported by the Israeli Smart Transportation 
Research Center (ISTRC).

 \bibliographystyle{splncs04}
 \bibliography{mybibliography} 
\end{document}